\documentclass[11pt,a4paper]{article}
\usepackage[hyperref]{emnlp-ijcnlp-2019}
\usepackage{times}
\usepackage{latexsym}
\usepackage{float}

\usepackage{times}
\usepackage{latexsym}
\usepackage{subfig}
\usepackage{comment}
\usepackage{times}
\usepackage{soul}
\usepackage{booktabs}
\usepackage{url}
\usepackage[utf8]{inputenc}
\usepackage{graphicx}
\usepackage{amsmath}
\usepackage{amssymb}
\usepackage{booktabs}
\usepackage{algorithm}
\usepackage{algorithmic}
\usepackage{multirow}
\usepackage{multicol}
\usepackage{caption}
\usepackage{xspace}

\usepackage{url}
\aclfinalcopy 

\title{Sequential Learning of Convolutional Features \\ for Effective Text Classification}

\author{Avinash Madasu \\
  Samsung R\&D Institute, Bangalore \\
  \texttt{m.avinash@samsung.com} \\\And
  Vijjini Anvesh Rao \\
  Samsung R\&D Institute, Bangalore \\
  \texttt{a.vijjini@samsung.com} \\}

\date{}

\begin{document}
\maketitle
\begin{abstract}
Text classification has been one of the major problems in natural language processing. With the advent of deep learning, convolutional neural network (CNN) has been a popular solution to this task. However, CNNs which were first proposed for images, face many crucial challenges in the context of text processing, namely in their elementary blocks: convolution filters and max pooling. These challenges have largely been overlooked by the most existing CNN models proposed for text classification. In this paper, we present an experimental study on the fundamental blocks of CNNs in text categorization. Based on this critique, we propose Sequential Convolutional Attentive Recurrent Network (SCARN). The proposed SCARN model utilizes both the advantages of recurrent and convolutional structures efficiently in comparison to previously proposed recurrent convolutional models. We test our model on different text classification datasets across tasks like sentiment analysis and question classification. Extensive experiments establish that SCARN outperforms other recurrent convolutional architectures with significantly less parameters. Furthermore, SCARN achieves better performance compared to equally large various deep CNN and LSTM architectures.
\end{abstract}

\section{Introduction}
Text classification is one of the major applications of Natural Language Processing (NLP). Text classification involves classifying a text segment into different predefined categories. Sentiment analysis of product reviews, language detection, topic classification of various news articles are some of the problem statements of text classification. Prior to the success of deep learning, text classification was dealt using lexicon based features. These primarily involve parts-of-speech (POS) tagging where verbs and adjectives are given more importance compared to other POS. Frequency based feature selection techniques like Bag-of-Words (BoW), Term frequency-Inverse document frequency (TF-IDF) have been used and the features are trained using machine learning classifiers like Logistic Regression (LR) or Naive Bayes (NB). These approaches provided strong baselines for text classification \cite{wang2012}.  However, sequential patterns and semantic structure between words play a crucial role in deciding the category of a text. Traditional lexicon approaches fail to capture such complexities. 
Since the success of deep learning, neural network architectures have outperformed traditional methods like BoW and count based feature selection techniques. Neural network architectures especially Convolutional Neural Networks (CNN) and Long Short Term Memory \cite{hochreiter1997long} (LSTM) have achieved state-of-the-art results in text classification. Shallow CNN architectures \cite{kim2014convolutional,madasu2019effectiveness,madasu2019gated} achieved admirable results on text classification as well. Very deep CNN architectures were proposed based on character level features \cite{zhang2015character} and word level features \cite{conneau2016very} which significantly improved the performance in text classification. 
In image classification and object detection, previously proposed deep CNN architectures \cite{szegedy2016rethinking} achieved state-of-the-art results. The effectiveness of deep CNNs can be very well explained where primary convolutional layers detect the edges of an object. As we go deeper, more complex features of an image are learnt. But in case of text, there hasn't been much understanding behind their success. Popular CNN architectures use convolution with a fixed window size over the words in a sentence. However, a question arises whether sequential information is preserved using convolution across words. \textit{What will be the effect if word sequences are randomly shuffled and convolution is applied on them?} In case of LSTM, random shuffling results in least performance as LSTMs rely on sequential learning and random ordering harms any such learning. Whereas, in case of a fixed window convolution applied across words, there is no strong evidence that it preserves sequential information. Previously proposed CNN architectures use max pooling operation across convolution outputs to capture most important features \cite{kim2014convolutional}. This leads to another question, if the feature selected by max pooling will always be the most important feature of the input or otherwise.

To answer the above questions, we conduct a study of experiments about the effects of using fixed window convolution across words as discussed in Section \ref{sec:filter}. And the effects of using max pooling operation on convolution outputs are discussed in Section \ref{sec:maxpool}.
Based on their critique, we propose a Sequential Convolutional Attentive Recurrent Network (SCARN) model for effective text classification. The proposed model relies on recurrent structures for preserving sequential information and convolution layers to learn task specific representations. In summary, the contributions of the paper are:
\begin{itemize} 
    \item We propose a new recurrent convolutional model for text classification by discussing the shortcomings of max pooling operation and also the strength and weakness of convolution operation.
    \item We evaluate the proposed model's performance on seven benchmark text classification datasets. The results show that the proposed model achieves better results with lesser number of parameters than other recurrent convolutional architectures.
\end{itemize}
\section{Related Work}

\subsection{Recurrent Convolutional Networks}
Previous architectures proposed for text classification were entirely convolution or recurrent based. There have been limited but popular works on combining recurrent structures with convolutional structures. \cite{lai2015recurrent} proposed a neural network (RCNN) model which combined recurrent and convolutional architectures. In their model each word is represented by a triplet of the word itself, its left and right context words, which are then trained sequentially using LSTM. Max pooling is then applied to capture the most important features of a sentence. Another attempt to combine recurrent and convolutional structures was made by \newcite{lee2016sequential} for sequential short-text classification. Their model uses LSTM to train short-texts and max pooling is applied on the outputs of all timesteps of LSTM to create a vector representation for each short-text. These architectures first use LSTM to train sequences and then max pooling operation is applied on its output. \newcite{zhou2015c} proposed a recurrent convolutional network (C-LSTM) where convolution is applied over windows of multiple pretrained word vectors (fixed window size) to obtain higher level representations. These features serve as input to an LSTM for learning sequential information.

\begin{figure*}%
   \centering
    \subfloat[Original Embeddings]{{\includegraphics[width=6cm]{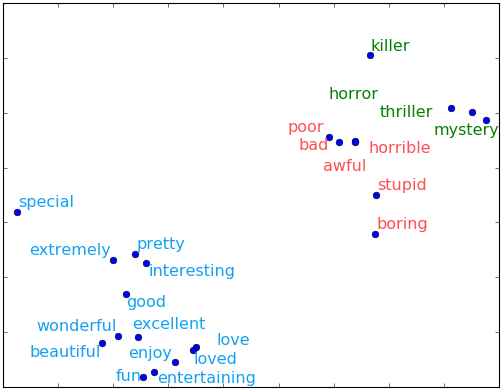}}}%
    \qquad
    \subfloat[Convolution Transformation]{{\includegraphics[width=6cm]{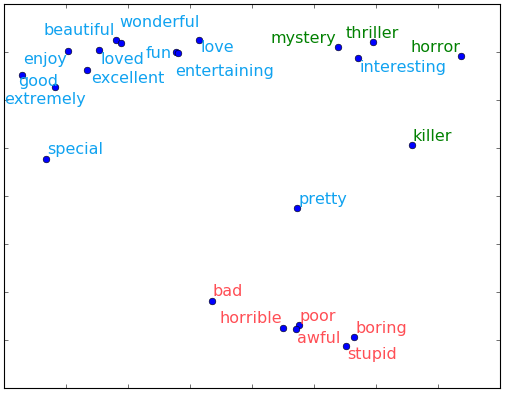}}}%
    \caption{t-SNE projection of original embeddings and after convolution transformation}%
  \qquad
    \label{fig:tsne}%
\end{figure*}

\subsection{Attention Mechanism}
Attention significantly improves the focusing on most meaningful information in the sentence. CNN and RNN architectures that were proposed with attention mechanism achieved superior performance to non-attention architectures. Attention originally proposed for machine translation \cite{bahdanau2014neural} was applied for document classification \cite{yang2016hierarchical} as well. Several variants of attention mechanism have been proposed like global and local attention \cite{luong2015effective} and self attention \cite{lin2017structured}. Attention has been applied to CNN blocks like pooling layers \cite{santos2016attentive} for discriminative model training. Recently, attention based architectures achieved state-of-art results in question answering \cite{devlin2018bert}.

\section{Understanding Convolution and Max pooling}
In this section we explore the working of CNNs in the context of natural language to better explain the intuition behind proposing SCARN. The key components of CNNs, namely convolution and max pooling strengths and weaknesses are discussed in this section. 

\subsection{Convolution operation}
\label{sec:filter}
The convolution operation in the context of its application on word embeddings is discussed as follows. 
\subsubsection*{What it doesn't learn}
Convolution operation can be explained as a weighted sum across timesteps. This may result in loss of sequential information. To illustrate our hypothesis, we conduct an experiment on the Rotten Tomatoes dataset \cite{Pang+Lee:05a}. In this experiment we train a CNN with convolution being applied on words with the fixed window size varying from one to maximum sentence length. We repeat this experiment with randomly shuffling all the words in an input sentence (Random ordering), and with shuffling every two consecutive words as shown in this example ``read book forget movie'' to ``book read movie forget'' (Alternate shuffle). The results of this experiment are illustrated in Figure \ref{fig:filters}\footnote{Experiment on more datasets could be found in the Appendix \ref{sec:Exp}}. Our observations in this experiment are the following:
\begin{itemize}
\item CNNs fail to fully incorporate sequential information as performance on random ordering and correct ordering are marginally near each other. As evident in the figure, the performance with correct ordering on window size 7 is comparable to performance with random ordering on window size 1.
\item As window size increases, the difference between correct and incorrect orderings diminishes and finally converges when window size is complete sentence length. This indicates that the ability to capture sequential information by CNNs, decline with increasing window size.
\end{itemize}
Furthermore, we also note that while performance on random ordering is marginally less than correct ordering, it is still higher than other context blind algorithms like bag-of-words as shown in Table \ref{tab:Accuracy}. This implies that while not fully exploiting sequential information, convolution filters still learn something valuable which we explore in next section.
\subsubsection*{What it learns}

To understand what convolution filters do, we train a CNN with single window size on the SST2 dataset \cite{socher2013recursive} for sentiment classification. As the convolution acts over a single word, it has no ability to capture sequential information. Hence, irrespective of input sentence word order, words will always have same respective convolution output. Essentially, convolution layer here acts like an embedding transformation, where input embedding space is transformed into another space. To understand this new space we project the respective  embeddings using t-SNE \cite{tsne} and illustrate them in Figure \ref{fig:tsne}. Here we make a distinction between the semantic knowledge of a word and the task specific knowledge of a word. As this experiment is done on dataset SST2, we refer to task specific knowledge as the sentiment knowledge of a word. Word embeddings\footnote{In this experiment GloVE word embeddings were used.} are generally trained on huge domain generic data because of which they capture the general semantic information of the word. Figure \ref{fig:tsne}a shows these word embeddings as they are in the original space. As we can see, semantically similar words cluster together. Figure \ref{fig:tsne}b shows words after their convolutional transformation. Blue coloured words represent words with positive sentiment. Red with negative sentiment and green words are those which are semantically close to negative sentiment words. However, in the context of movie reviews, their sentiment value is close to positive sentiment words. From this experiment we observe:
\begin{itemize}
    \item The transformation of the green words between original and convoluted outputs, shows that the convolution layer is able to tune their input embeddings such that they are more closer to the positive sentiment cluster than to their original semantic cluster. For example, a word like ``killer'' in the original embedding  space is very close to negative sentiment words like, ``bad'' and ``awful'', especially if the word embeddings were produced on huge fact based datasets like Wikipedia or news datasets. However in the SST2 dataset, ``killer'' is often used to describe a movie very positively. Hence sequential learning on such transformed embeddings will be more effective than the original semantic embeddings.
    \item Some words like ``pretty'' might be semantically closer to the positive sentiment words. However in the dataset, we find common occurrences of phrases like ``pretty good'' or ``pretty bad''. The layer has hence transformed its position nearly equidistant to both the negative and positive clusters. In other words, convolution layer is able to robustly create a transformation which can handle any task specific noise.
\end{itemize}

\begin{figure}%
\centering
\includegraphics[width=7cm]{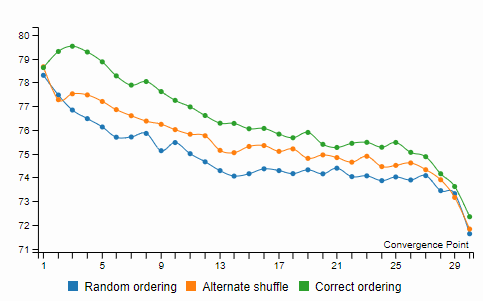}
\caption{Accuracy (y-axis) percentage on Rotten Tomatoes dataset with varying window size.}
\label{fig:filters}
\end{figure}

The above observations show the effectiveness of the convolution operation. This is because a single convolution filter output value captures a weighted summation over all the features of the original word embedding. This enables the network to learn more task appropriate features as many as the number of filters. 

With the above points in mind, we observe that convolution operation does not fully exploit sequential information, especially on larger window sizes. However we do see their effectiveness in conforming the input embeddings to a representation specifically meaningful to the task. Hence, for sequential learning instead of relying on convolution, we use recurrent architectures in our proposed SCARN model.


\begin{figure}%
\centering
\includegraphics[width=7cm]{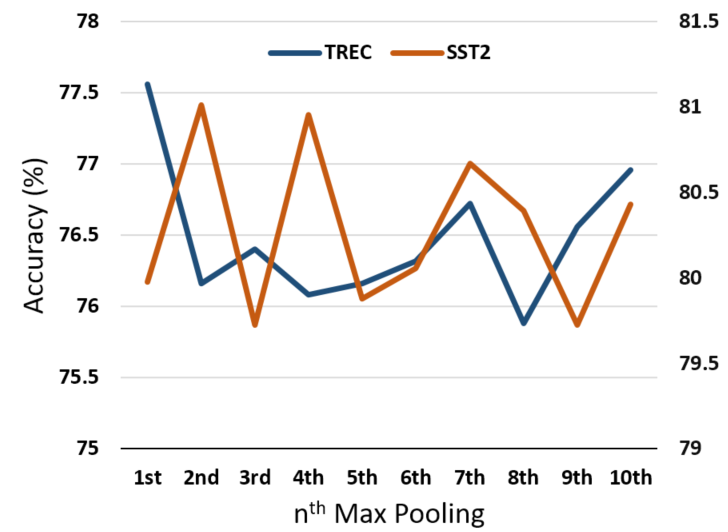}
\caption{Accuracy (y-axis) percentage on TREC and SST2 datasets with varying \textit{n} for the \textit{$n^{th}$} Max pooling.}
\label{fig:rt10}
\end{figure}

\subsection{Max pooling operation: More doesn't mean better}
\label{sec:maxpool}
Max pooling operation in images identify the discriminative features from convolution outputs. However this does not relate the same way of selecting most relevant features from convoluted features in texts. To illustrate this, we perform an experiment using a CNN architecture based on architecture by \newcite{kim2014convolutional} popularly used in text classification. We define \textit{$n^{th}$} Max pooling as choosing $n^{th}$ highest value of all filters as opposed to first ($n$=$1$). By analyzing the performance distinctions on varying values of n, we make an attempt to assess whether maximum necessarily means task meaningful. The results are illustrated in Figure \ref{fig:rt10}\footnote{Experiment on more datasets could be found in the Appendix \ref{sec:Exp}}. The results show that when $n$ is increased the performance varies arbitrarily on all datasets. This shows that there is no apparent co-relation between magnitude of the values to importance for the task. Hence, this relationship does not stand strong for text based inputs.
\section{Model}
\subsection{Overview}
\begin{figure}%
\includegraphics[width=8cm]{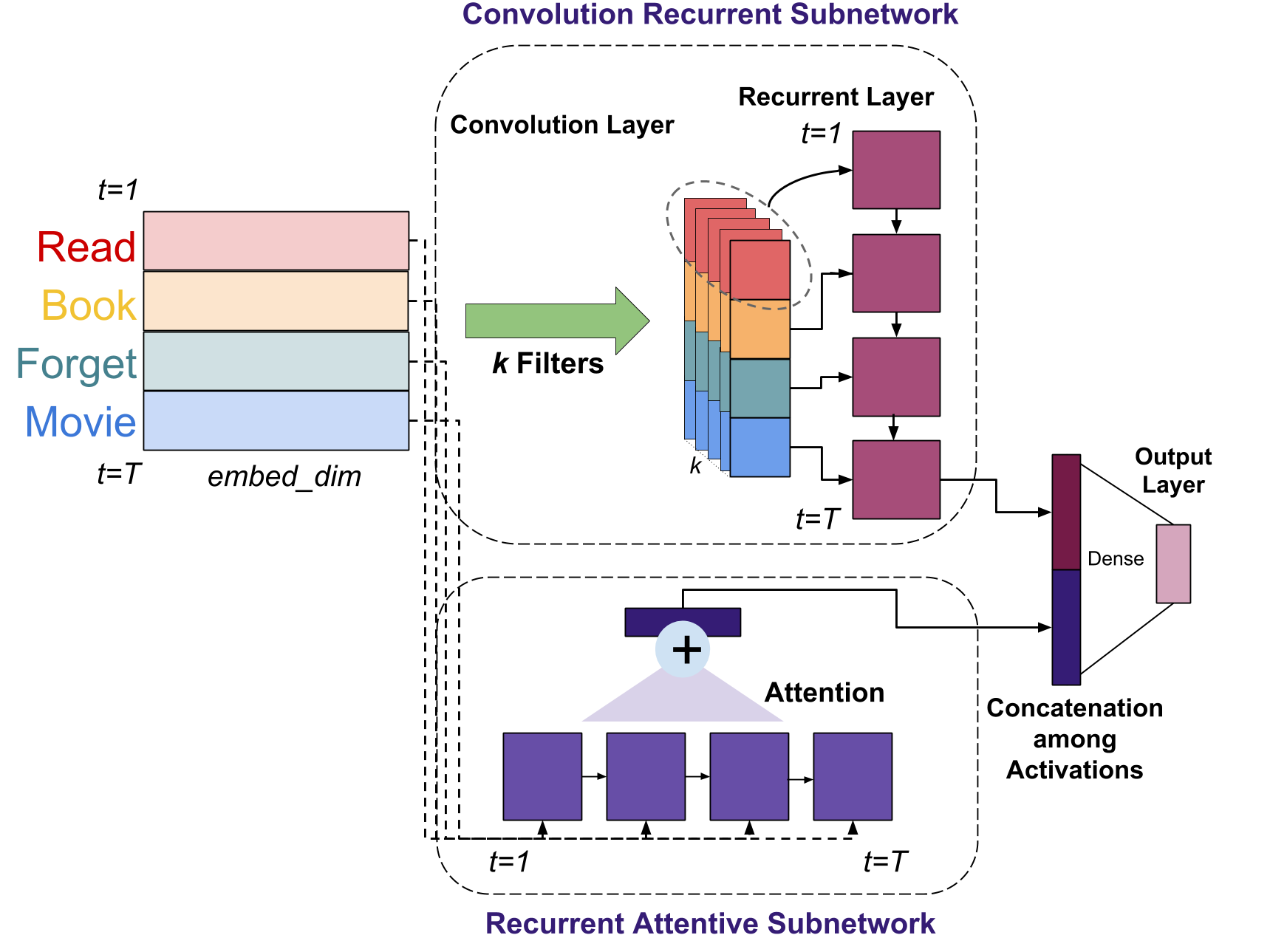}
\caption{SCARN Architecture}
\label{fig:scarn}
\end{figure}

The proposed SCARN model architecture is shown in the Figure \ref{fig:scarn}. 
Let $V$ be the vocabulary size considered and X $\in$ $\mathbb{R}^{V \times d}$ represent embedding matrix where $X_{i}$ is a $d$ dimensional word vector. Vocabulary words contained in the pretrained embeddings are initialized to their respective embedding vectors and words that are not present are assigned 0's. For each input text, a maximum length $N$ is considered. Zero padding is applied if the length of input text is less than $N$. Hence, each sentence or paragraph is converted to I $\in$ $\mathbb{R}^{N \times d}$ dimensional vector which is the input to SCARN model.

The model consists of two subnetworks: Convolution Recurrent subnetwork and Recurrent Attentive subnetwork. In the first subnetwork, convolution  with single window size is applied on the input $I$. Convolution filters learn the higher level representations from the input text. The output from convolution is trained sequentially using LSTM network. In the second subnetwork, the input $I$ is trained using LSTM. To better focus on most relevant words, attention mechanism is applied to the outputs of LSTM from every time step. Attention creates an alternate context vector for the input $I$ by choosing the most suitable words necessary for classification. The outputs from the first subnetwork and second subnetwork are concatenated and connected to the output layer through a dense connection.

\subsection{Convolution Recurrent subnetwork}
Let $I$ be represented as sequence of words $I$ = $w_{1}w_{2}w_{3}...w_{N}$ where $w_{i}$ represent the $i^{th}$ word in the input. A total of $K$ convolution filters are applied on each word $w_{i}$ of input. Let $L$ $\in$  $\mathbb{R}^{K \times d}$ be the weight
matrices of all filters. For each filter $l$ $\in$ $\mathbb{R}^{1 \times d}$ when applied on  $w_{i}$, outputs a new feature $C_{il}$. Therefore, a new feature vector $C_{i}$ is obtained for each word $w_{i}$ after convolving  with $K$ filters.
\begin{equation}
     C_{i} = C_{i1}C_{i2}C_{i3}......C_{iK}
\end{equation}
Similarly, the above procedure is repeated for all the words in the input to produce a feature vector $C$ $\in$ $\mathbb{R}^{N \times d \times K}$.
\begin{equation}
     C = f(I \circledast L)
\end{equation}
where $\circledast$ represents convolution operation and $f$ is the non-linear activation (ReLU). Applying convolution to individual words preserve the sequential information. Convolution learns the higher level representations for the input words. Each word will be transformed to a new representation pertinent to the task. The new feature representations are trained sequentially using LSTM. 

\subsection{Recurrent Attentive subnetwork}
In this subnetwork, the input word embeddings are trained using LSTM. All words in the input sentence are not equally important in predicting final output.  Although, LSTM learns sequential information, selecting significant information is a key issue for text classification. For this purpose, we employ an attention layer on the top of LSTM outputs. Attention mechanism focuses on specific significant words and tries to construct alternate context vector by aggregating the word representations.

\begin{figure}%
\centering
\includegraphics[width=6cm]{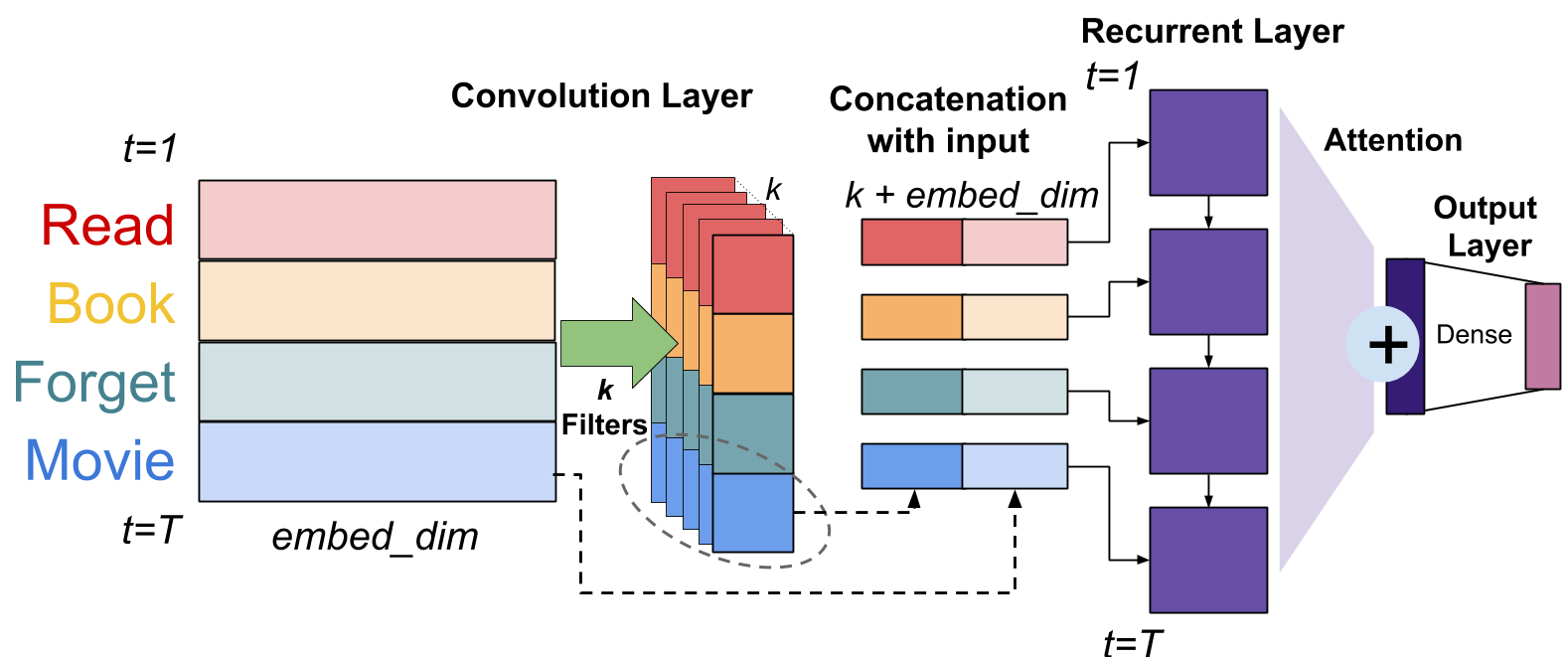}
\caption{concat-SCARN architecture}
\label{fig:concanarch}
\end{figure}
\section{Experiments}

\subsection{Datasets}
We tested our model on standard benchmark datasets: Rotten-Tomatoes (RT) \cite{Pang+Lee:05a}, Polarity V2.0 (Pol) \cite{Pang+Lee:04a}, SST-2 \cite{socher2013recursive}, Home and Kitchen reviews (AR) \cite{he2016ups}, TREC \cite{Li:2002:LQC:1072228.1072378}, IMDB \cite{maas-EtAl:2011:ACL-HLT2011} and  Subjectivity Objectivity (SO) \cite{Pang+Lee:04a}. The statistics for the datasets are shown in Table \ref{tab:Statistics}

\begin{table*}[t]
  \centering
  \setlength{\tabcolsep}{4pt}
  \renewcommand{\arraystretch}{1.25}
  \begin{tabular}{{l l c c c c c c c}}
    \hline
    & Model & IMDB & TREC & SO & RT & Pol & AR & SST-2 \\
    \hline
    \hline
    \multirow{2}{*}{Linear} &  BoW + LR & 86.452 & 73.600 & 82.800 & 64.320 & 76.500 & 47.000 & 80.500 \\
     & TFIDF + LR & 78.740 & 72.000 & 84.000 & 63.237 & 77.250 & 49.200 & 80.340 \\
    \hline
    \multirow{2}{*}{Word Vector} & Avg Word Vectors + MLP & 85.44 & 86.999 & 90 & 75.691 & 68.25 & 47.015 & 81.219 \\
     & Paragraph2Vec + MLP & 77.472 & 45 & 77.2 & 62.996 & 77.25 & 38.925 & 67.27\\
    \hline
    \multirow{2}{*}{CNN} & Deep CNN  & 78.44 & 28.999& 85.799 & 75.4 & 69.999 & 40.333 & 64.305 \\
    & Char CNN & 70.484 & 74.199 & 59 & 50 & 50.249 & 34.37 & 60.516 \\
    \hline
     \multirow{2}{*}{RNN} & LSTM & 88.37 & 76.99 & 90.899 & 77.436 & 74.5 & 47.85 & 78.747 \\
     & Bi-LSTM  & 88.69 & 75.8 & 91.399 & 76.594 & 77.499 & 51.569 & 79.242 \\
    \hline
     \multirow{2}{*}{Attention} & LSTM+Attention  & 88.35 & 77 & 89.7 & 76.895 & 77.75 & 50.568 & 80.01\\
      & concat-SCARN & 88.88 & 78.8 & 89.8 & 77.858 & 75.75 & 51.569 & 80.06 \\
    \hline
    \multirow{2}{*}{RNN-CNN} & RCNN  & 86.607 & 79.6 & 91.1 & 78.098 & 78.25 & 48.026 & 80.395 \\
     & C-LSTM & 87.676 & 90.4 & 91.3 & 76.474 & 67 & 52.784 & 78.308 \\
    \hline
    \hline
     Our model & SCARN & \bf 89.788 & \bf 90.799 & \bf 92.4 & \bf 79.641 & \bf 78.75& \bf 53.350 & \bf 82.262 \\
    \hline
  \end{tabular}
  \caption{Accuracy scores in percentage of all models on every dataset}
  \label{tab:Accuracy}
\end{table*}

\begin{table}[]
    \centering
    \begin{tabular}{cc}
    \hline
        Model & No. of parameters  \\
        \hline
        \hline
         Small SCARN &68,425 \\
         Large SCARN& 166,639\\
         RCNN & 180,601 \\
         C-LSTM & 676,651 \\
         \hline
    \end{tabular}
    \caption{Number of parameters for each model}
    \label{tab:parameters}
\end{table}

\begin{table*}[]
    \centering
    \setlength{\tabcolsep}{8pt}
    \begin{tabular}{c c c c c c c}
    \hline
        Dataset & Dataset size & Train & Dev & Test & Max Vocab size & Classes  \\
        \hline
        \hline
         IMDB & 50000 & 20000 & 5000 & 25000 & 30000 & 2 \\
         TREC & 5952 &4906 &546  &500 & 5000&6 \\
         SO &10000 &8100 &900 &1000 & 30000&2\\
         RT & 10662 &8100 & 900 &1662 & 30000&2\\
         Pol & 2000 & 1280&320 & 400&30000 & 2\\
         AR & 121565 &80000 &20000 &21565 &30000 &5\\
         SST-2 & 9613 &6920 & 872 & 1821 & 10000 & 2\\
         \hline
    \end{tabular}
    \caption{Summary Statistics of all datasets}
    \label{tab:Statistics}
\end{table*}

\subsection{Baselines}
We compared our model to various text classification approaches\footnote{For lack of code, results are from our implementations}.
\subsubsection*{BoW and TF-IDF + LR}
Bag-of-words(BoW) and TF-IDF are strong baselines for text classification. BoW and TF-IDF features are extracted and softmax is applied on the top for classification.
\subsubsection*{Average Word Vectors + MLP}
This baseline uses the average of word embeddings as features for the input text which are then trained using Multilayer perceptron (MLP).
\subsubsection*{Paragraph2Vec + MLP}
Each input sentence is converted to a feature vector using Paragraph2vec\cite{le2014distributed} which are then trained using Multilayer perceptron (MLP).
\subsubsection*{Deep CNN}
This baseline is based on deep CNN architecture \cite{conneau2016very}  with approximately matching number of parameters as our model.
\subsubsection*{Char CNN}
For our comparison, we employed the Char CNN architecture \cite{zhou2015c} with less parameters compared to the original model, as the datasets used by them were considerably huge than ours.
\subsubsection*{LSTM and Bi-LSTM}
We also offer a comparison with LSTM and Bi-LSTM architectures with a single hidden layer and approximately matching number of parameters as our model.
\subsubsection*{LSTM + Attention}
In this baseline, attention mechanism is applied on the top of LSTM outputs across different time steps.
\subsubsection*{concat-SCARN}
In this model, we concatenate the outputs from convolution layer as in SCARN, with input word embeddings at each time step. The concatenated outputs are trained using LSTM. Attention is applied on the top of this layer. Figure \ref{fig:concanarch} shows the architecture of concat-SCARN model.
\subsubsection*{RCNN}
We compared our model to the RCNN model \cite{lai2015recurrent} which uses max pooling for selecting the most appropriate features of a sentence.

\begin{figure*}
\centering
\includegraphics[scale=0.5]{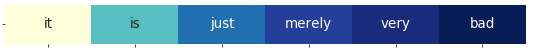}
\includegraphics[scale=0.5]{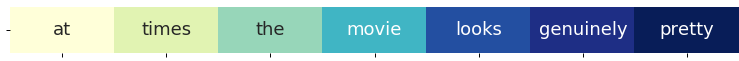}
\includegraphics[scale=0.5]{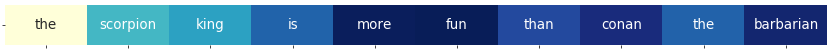}
\caption{Attention weights for some of the sentences from the SST2 dataset}
\label{fig:atten}
\end{figure*}

\subsubsection*{C-LSTM}
We compared our model to C-LSTM model \cite{zhou2015c} which uses convolution over fixed window of words to learn higher level representations.
\subsection{Implementation}
\subsubsection*{Input and Training Details}
We used google pretrained word vectors for C-LSTM\footnote{https://code.google.com/archive/p/word2vec/}, as it was used in the original work. For all the other experiments which require embeddings, we use GloVe pretrained word vectors\footnote{https://nlp.stanford.edu/data/glove.840B.300d.zip}. The size of a word embedding in this model is 300. For each dataset, a maximum sentence length is considered which is 30 for TREC, SO, RT, Pol, SST-2 datasets, 400 for IMDB and 100 for AR dataset.

We apply a dropout layer \cite{srivastava2014dropout} with a probability of 0.5 on the pretrained embeddings. We also apply dropout with a probability 0.5 on the dense layer that connects to output. We use Adam as the optimizer with a batch-size of 16 for small SCARN model and 50 for Large SCARN model. The initial learning rate is set to 0.0003. Training is done for 30 epochs.

\subsubsection*{Architecture}
We employ two different architectures of SCARN model since the datasets vary in size. For datasets TREC, SO, RT, Pol, SST-2 the number of convolution filters are 50. Number of LSTM cells considered for these datasets are 32. We call this architecture \textbf{Small SCARN model}. For datasets AR and IMDB, number of convolution filters are 100 and number of LSTM cells are 64. We call this architecture \textbf{Large SCARN model}. The comparison of number of parameters for each model is shown in Table \ref{tab:parameters}. ReLU is used as the activation function in convolution layers. In the output layer, softmax is used for multi-class classification and sigmoid for binary.

\section{Results and Discussion}
Results of the experiments are tabulated in Table \ref{tab:Accuracy}. We observe that, 
the proposed SCARN model outperforms linear and word vector based models, because of their inability to incorporate sequential information. We also compare our model to recurrent models like LSTM, Bidirectional-LSTM and find that SCARN outperforms them as these recurrent architectures even though learn sequential information, lack SCARN's learning of task specific representations through convolution. Meanwhile, SCARN outperforms deep CNN and char CNN models, for their lack of learning sequential information the same way recurrent architectures can.
When SCARN is compared to other recurrent convolutional architectures RCNN and C-LSTM, SCARN achieves significantly much better performance with lesser parameters as shown in Table \ref{tab:parameters}. RCNN uses max pooling to capture the most important feature. However our discussions in Section \ref{sec:maxpool} show max pooling's choice of maximum may not necessitate importance. C-LSTM used fixed window convolution across words. But as seen in Section \ref{sec:filter} fixed window convolution do not capture sequential information adequately. In SCARN model, we apply convolution across single word to learn task specific information, on which LSTM architecture is trained to capture contextual information.

\begin{figure}%
    \centering
    \subfloat[Mean]{\includegraphics[width=7cm]{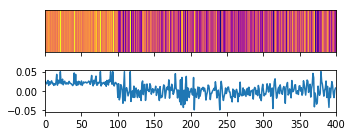}}%
    \qquad
    \subfloat[Standard Deviation]{\includegraphics[width=7cm]{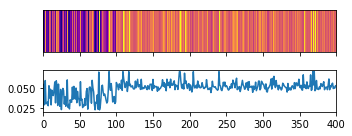}}%
    \caption{Statistics of each feature in concatenation layer outputs on the IMDB dataset's training set.}
    \label{fig:concatheatmap}%
\end{figure}

We observe that SCARN significantly outperforms concat-SCARN on all the datasets. We believe this is because in concat-SCARN, concatenation between input embeddings and convolution activations is done. However they belong to a very different distribution. In Figure \ref{fig:concatheatmap}, we show the average and standard deviation of outputs at the concatenation layer of concat-SCARN on IMDB dataset's training set. As evident from the figure, we can make a stark distinction between the two concatenated sections as shown in the architecture of concat-SCARN in Figure \ref{fig:concanarch}. Because of the irregular distributions, the learning may become biased towards one of the segments. Furthermore, attempts to address this issue by using batch normalization layers may not be a good idea, as normalizing word embedding inputs may spoil the semantic concepts underlying them. Hence, in our SCARN model, subnetworks are trained differently, and final activations are concatenated as illustrated in SCARN's architecture in Figure \ref{fig:scarn}. As long as both subnetworks' outputs come from similar activation function, they will have similar distributions as well and hence they won't suffer from the same problem as concat-SCARN. 

Effectiveness of attention has also  been illustrated in Figure \ref{fig:atten}, where attention weights have been shown in a heat map, with darker colors corresponding to a higher weightage. We observe that SCARN is able to effectively utilize attention weights to focus on most contributing input words.

\section{Conclusion}
In this paper we present a critical study and viewpoint of CNNs, which even though are popularly used in text classification, details of it are often overlooked. 
We find that convolutional filters learn particularly in the context of sequential information. But at the same time, they are good at learning higher level task-relevant features. On the other hand, we find max pooling to be very arbitrary in selection of crucial features and hence contributing minimal to the overall task. We also find that the problems with input concatenation, as it imbalances the representations because of difference in nature of distributions. Based on our study we proposed SCARN, for effectively utilizing convolution and recurrent features for text classification. Our model beats other popular ways of combining recurrent and convolutional architectures with quite less number of parameters on various benchmark datasets. Our model also outperforms CNN and RNN architectures with equally deep or same number of parameters.

\bibliography{emnlp-ijcnlp-2019}
\bibliographystyle{acl_natbib}

\appendix

\begin{figure*}[t]%
   \centering
    \subfloat[RT]{{\includegraphics[width=4.5cm]{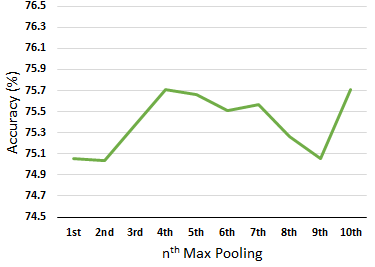}}}%
    \subfloat[IMDB]{{\includegraphics[width=4.5cm]{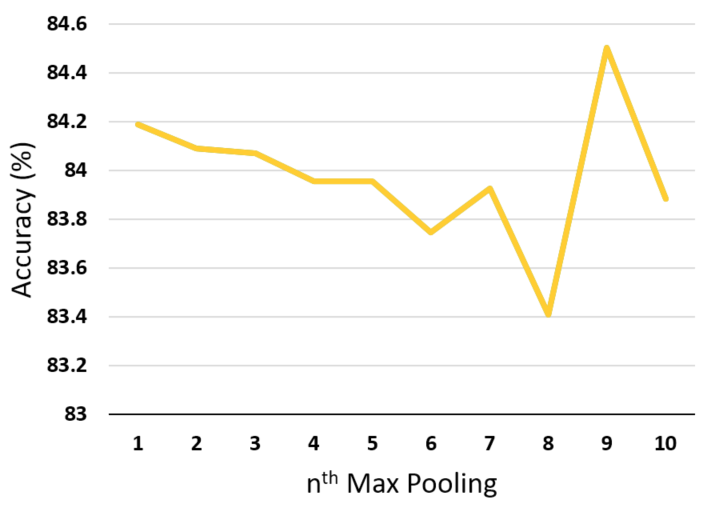}}}%
    \subfloat[SO]{{\includegraphics[width=4.5cm]{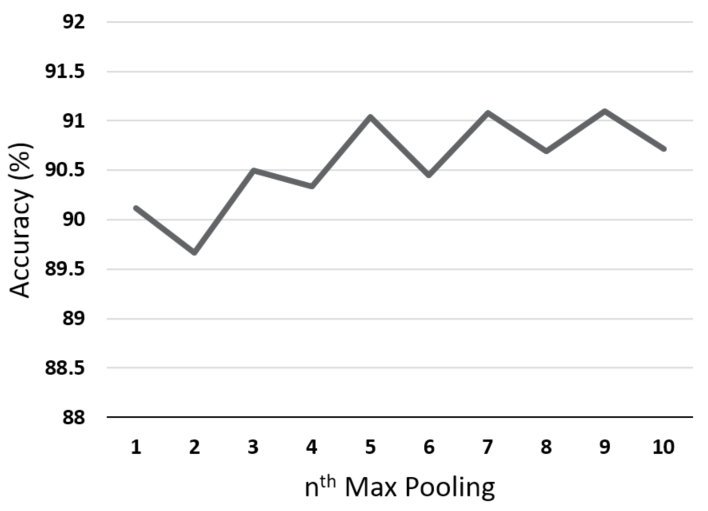}}}%
    \caption{\textit{$n^{th}$} Max pooling experiments on RT, IMDB and SO Datasets}
    \label{fig:maxpool}%
\end{figure*}

\begin{figure}%
   \centering
    \subfloat{{\includegraphics[width=2cm]{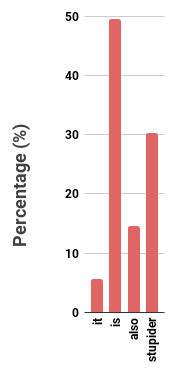}}}%
    \qquad
    \subfloat{{\includegraphics[width=4cm]{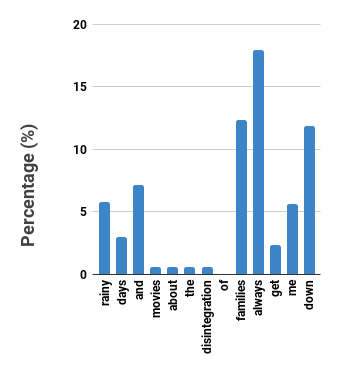}}}%
    \caption{Percentage distribution of max pooling outputs for misclassified samples from SST2 dataset}
  \qquad
    \label{fig:maxpoolPerc}%
\end{figure}

\section{Max Pooling: missclassified examples}
\label{sec:mp10}
For this experiment, after convolution over single word, we perform max pooling across filter outputs and attempt to discern the max pooling outputs. We identify which input words' convolution outputs were chosen by the max pooling operation. These distributions illustrated in Figure \ref{fig:maxpoolPerc} are presented on some misclassified examples from the SST2 dataset. We see that often words that should have been contributing most to the overall sentiment value, have very less or minimal share. For example in the sentence, ``it is also stupider'', we see ``is'' having the near majority share, even though it tells nothing about the sentiment of the sentence. At the same time in ``rainy days and movies about the disintegration of families always get me down'', ``disintegration'' has almost no share despite being an important word for evaluating sentiment.

\section{Experiments on more datasets} 
\label{sec:Exp}
\begin{figure}%
\centering
\includegraphics[width=7cm]{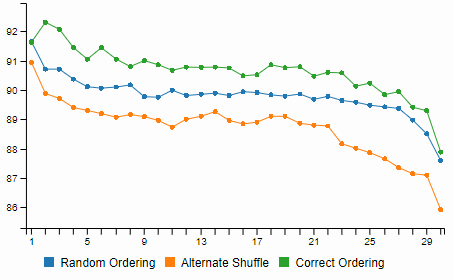}
\caption{Accuracy (y-axis) percentage on SO dataset with varying window size.}
\label{fig:SOfilters}
\end{figure}

\begin{figure}%
\centering
\includegraphics[width=7cm]{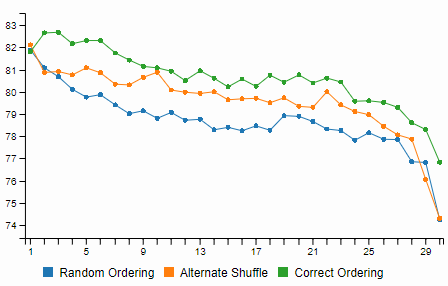}
\caption{Accuracy (y-axis) percentage on SST2 dataset with varying window size.}
\label{fig:SSTfilters}
\end{figure}

Figure \ref{fig:maxpool} shows the \textit{$n^{th}$} Max pooling experiment explained in Section \ref{sec:maxpool} on other datasets, namely RT, IMDB and SO. We still find that there is no co-relation between importance for task and magnitude of the values. Figure \ref{fig:SOfilters} and Figure \ref{fig:SSTfilters} show the Ordering experiment from Section \ref{sec:filter} on SO and SST2 datasets respectively. As evident from the figures, the proposed hypotheses stay sound on other datasets as well.


\end{document}